# Mitigating Attrition: Data-Driven Approach Using Machine Learning and Data Engineering

**Naveen Edapurath Vijayan**

Sr Data Scientist, Amazon, Seattle, WA 98765
nvvijaya@amazon.com

**Abstract**
This paper presents a novel data-driven approach to mitigating employee attrition using machine learning and data engineering techniques. The proposed framework integrates data from various human resources systems and leverages advanced feature engineering to capture a comprehensive set of factors influencing attrition. The study outlines a robust modeling approach that addresses challenges such as imbalanced datasets, categorical data handling, and model interpretation. The methodology includes careful consideration of training and testing strategies, baseline model establishment, and the development of calibrated predictive models. The research emphasizes the importance of model interpretation using techniques like SHAP values to provide actionable insights for organizations. Key design choices in algorithm selection, hyperparameter tuning, and probability calibration are discussed. This approach enables organizations to proactively identify attrition risks and develop targeted retention strategies, ultimately reducing costs associated with employee turnover and maintaining a competitive edge in talent management.

## 1. INTRODUCTION

Employee attrition, or the voluntary departure of employees from an organization, is a significant challenge faced by companies across various industries. High attrition rates can lead to substantial costs associated with recruiting, hiring, and training new employees, as well as disruptions in productivity and knowledge transfer. Moreover, the loss of valuable talent can hinder an organization's ability to remain competitive and achieve its strategic objectives.

Traditional approaches to mitigating attrition often rely on reactive measures, such as exit interviews and employee satisfaction surveys. However, these methods may fail to capture the complex interplay of factors that contribute to an employee's decision to leave. Additionally, they do not provide a proactive and data-driven approach to identifying and addressing attrition risks before they manifest.

In recent years, the advent of machine learning and data engineering techniques has opened new avenues for organizations to tackle complex business problems, including employee attrition. By leveraging the power of data and advanced analytical methods, companies can gain valuable insights into the drivers of attrition and develop targeted strategies to retain their workforce.

This paper presents a novel approach to mitigating employee attrition by combining machine learning algorithms and data engineering practices. A comprehensive framework is proposed that integrates data from various sources, such as human resources information systems, employee surveys, and organizational performance metrics. Through data preprocessing, feature engineering, and the application of machine learning models, the approach aims to identify patterns and predictors of attrition, enabling organizations to take proactive measures to retain their valuable employees.

## 2. DATA PREPARATION & FEATURE ENGINEERING

To build an effective model for predicting and mitigating employee attrition, it is crucial to leverage multiple data sources from various human resources systems. The goal should be to generate a comprehensive set of features that capture a wide array of information about employees, their teams, and their managers. Depending on the organization, the framework could potentially incorporate hundreds of features for each employee.

Feature engineering will play a critical role in this process, and several key considerations should be taken into account:

- **Proxy Features for Attrition Drivers: Many** features should be engineered to create proxies that support hypotheses around the drivers of attrition. These features could cover various subject areas, such as employee demographics, organizational churn, compensation, performance, feedback and alignment, talent movement velocity, employee sentiment, and manager and team characteristics.
- **Consistency and Reliability:** Ensure that features are engineered from consistent and reliable data sources.
- **Time-Sensitivity:** Engineer features in a way that enables back-testing and prevents information leakage from the future. This will allow for proper evaluation of the model on historical data.
- **Missing Value Handling:** Develop extensive logic to impute missing feature values based on the nature of the feature and the prevalence of missing data.
- **Automated Feature Engineering:** In addition to manual feature engineering, explore automated techniques such as driverless AI and machine learning frameworks designed to handle missing values. These can help create compact representations of input features.

The feature engineering process should aim to capture a comprehensive set of factors that may influence employee attrition, while ensuring data quality, consistency, and interpretability. Additionally, it is important to consider the ability to engineer features as of a specific date, allowing for proper model validation and testing.

By carefully considering these ideas and approaches during the data preparation and feature engineering stages, organizations can lay a solid foundation for building an effective model to predict and mitigate employee attrition, leveraging the available employee data and insights.

## 3. MODELING APPROACH

This section outlines the considerations and ideas for selecting training and test datasets, establishing a baseline model, developing the proposed model, evaluating performance, interpreting the model, and generating model outputs.

**A. Training and Testing Strategy**

- Collect at least an year's worth of historical data from various HR systems and data sources for cleaning

and feature engineering.
- Engineer features as of the end of each week/month over that year, capturing the state of each employee and their characteristics at those specific points in time.
- Construct a panel data structure that starts 12 months before the prediction month and ends X months before the prediction month, where X is the label horizon (e.g., 3 months for predicting 3-month attrition).
- This panel data structure ensures capturing the cyclical nature of attrition patterns, important data cuts by which attrition varies significantly, and preventing any employee overlap between train, validation, and test datasets.
- Encode a binary attrition label (0 or 1) indicating whether an employee stayed or left the organization within the prediction horizon.
- Split the data into training (e.g., 75%), validation (e.g., 15%), and test sets, stratified by relevant employee demographics (e.g., job family, job level, location) and the attrition label to maintain class balance.
- Employ a custom train-test split technique that samples data at the employee level and stratifies for the most important employee characteristics, ensuring no employee overlap between datasets.
- Use the validation set for early stopping the training process based on a custom evaluation metric (e.g., AUC-PR) and for calibrating the model's output probabilities.

**B. Baseline and Proposed Models**
- Establish a baseline model using basic features (e.g., job level, job family, tenure, location, quarter of the year) and a machine learning algorithm like an isotonic calibrated gradient boosting model (e.g., LightGBM).
- Train models to predict various outcomes of interest (e.g., regretted attrition, unregretted attrition, total attrition, transfers, total movement) over multiple time horizons (e.g., 1 month, 3 months, 6 months, 12 months).

**C. Model Calibration**
- Address class imbalance in the target variable (attrition) by using techniques like cost-sensitive learning or down sampling the majority class during training.
- Calibrate the model's output probabilities using techniques like isotonic regression to adjust for biases introduced by class imbalance or the use of non-probabilistic algorithms like decision trees or ensembles. Techniques like isotonic regression maps the model's output to the right empirical ratio values without changing the probability score order of the instances, ensuring that the estimated class probabilities reflect the true underlying probability of the sample.

**D. Evaluation Metrics and Model Interpretation**
- Utilize appropriate evaluation metrics to assess model performance, such as AUC-PR (Area Under the Precision-Recall Curve) for imbalanced datasets, precision, recall, F1-score, and mean absolute percentage error for probability predictions.
- Employ techniques like feature importance analysis, partial dependence plots, and SHAP (SHapley Additive exPlanations) values to interpret the model and understand the key drivers of attrition.
- Analyze the model's performance and feature importances across different employee segments, locations, and time horizons to gain deeper insights.

### E. Model Output

- Generate individual-level predictions of attrition probability for each employee within the specified time
- horizon.
- Aggregate predictions by relevant data cuts (e.g., teams, departments, locations, job families) to provide insights and attrition risk scores to stakeholders and business leaders.
- Develop visualizations, dashboards, and reports to communicate model outputs, insights, and actionable recommendations effectively to various stakeholders.

By considering these ideas and approaches, organizations can develop a robust and interpretable model for predicting and mitigating employee attrition, tailored to their specific data and requirements. The model can provide valuable insights to inform talent retention strategies, resource allocation, and proactive interventions to address attrition risks.

## 4. FEATURE IMPACT AND MODEL INTERPRETATION

Examining feature impact and interpreting the model is crucial for several reasons. First, it helps identify promising areas for improving model accuracy by building hypotheses around influential features. Second, it provides insights into how and why certain features matter, serving as a starting point for prioritizing root cause analyses. Third, it demonstrates the value added by specific data sources or feature sets, such as employee sentiment data.

To understand feature importance, visualizations can be created to show the top features by their relative importance scores in the model. However, these importance scores alone do not reveal how the features influence the predicted probabilities. To address this, techniques like SHAP (SHapley Additive exPlanations) can be employed to visualize feature contributions to the predicted probabilities. By analyzing these visualizations, insights can be gained into how specific features influence the model's predictions. For example, features capturing employee sentiment or engagement may have a negative impact on attrition probability when their values are high, indicating that employees with higher engagement scores are less likely to leave the organization.

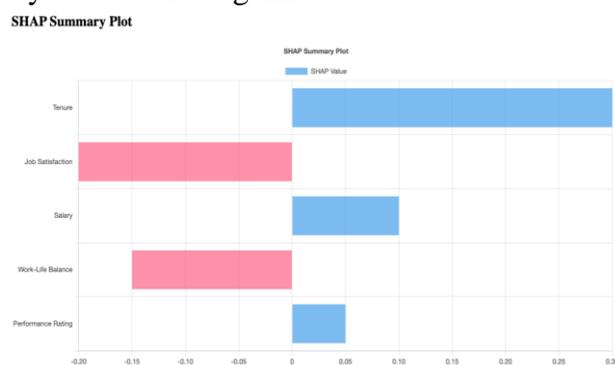

Additionally, features capturing opposing sentiments or characteristics (e.g., favorable vs. unfavorable answer ratios) can be expected to have contrasting effects on the predicted probabilities, as demonstrated by their positions on the SHAP value plot. These visualizations and interpretations serve as a starting point for deeper investigations and root cause analyses. Organizations can leverage these insights to prioritize areas for improvement, develop targeted interventions, and ultimately enhance their talent retention strategies. Furthermore, by comparing the feature importance and contributions across different model variants (e.g., with and without specific data sources), the value added by certain feature sets can be quantified and evaluated. To facilitate broader adoption and understanding of these models, organizations can consider developing toolkits or resources for the HR community, enabling them to interpret and explain their own models effectively.

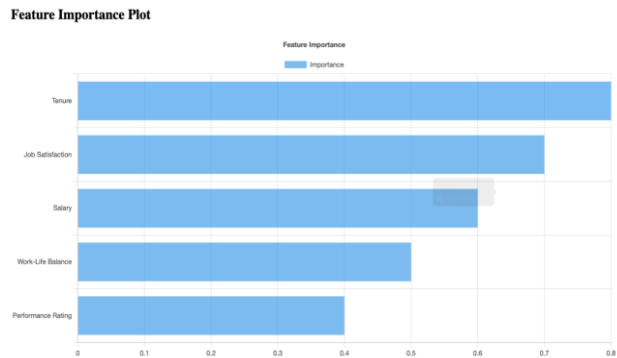

## 5. DESIGN CHOICES

This section presents the design choices that needed to be made, lists the considered solutions, and discusses the final choice made to address each challenge.

### a. Imbalanced Dataset

Given the typically low attrition rates, the dataset is highly imbalanced, with a significant disparity between the positive (attrition) and negative (retention) classes. The following approaches were considered:

- Resampling: Up-sampling the positive class, down-sampling the negative class, and a combination of both.
- Synthetic Sample Generation: Using algorithms like SMOTE (Synthetic Minority Over-sampling Technique) to generate synthetic samples.
- Cost-Sensitive Learning: Penalizing the classifier by imposing an additional cost on the model for misclassifying the minority class during training.
- Different Algorithms: Trying different implementations of neural networks, linear models, and decision trees to assess their ability to handle imbalanced datasets.

### b. Categorical Data:

Several features are categorical in nature, such as job family, location, and job indicators. To handle categorical data, the following techniques can be considered:

- One-Hot Encoding: Encoding each category as a dummy variable.
- Encoding as Ordinal: Converting categorical features to ordinal features by assigning arbitrary orders.
- Feature Hashing: Applying a hashing function to the categories and representing them by their indices.
- Target Encoding: Replacing categorical values with the mean of the target variable for that category.
- Neural Network Embeddings: Creating embedding layers to project categorical features from a high-dimensional sparse space to a low-dimensional dense space.

### c. Evaluation Metric:

Various evaluation metrics can be considered, including accuracy, precision, recall, F1-score, AUC (Area Under the Receiver Operating Characteristic Curve), and AUC-PR (Area Under the Precision-Recall Curve).

### d. Machine Learning Algorithms:

Different algorithms and implementations can be explored, including linear models (logistic regression, SVM), decision trees (Random Forest, XGBoost, LightGBM, CatBoost), neural networks (feed-forward neural networks), and TabNet.

*e. Hyperparameter Tuning:*

LightGBM has a wide range of hyperparameters that can be tuned for faster speed, better accuracy, and overfitting prevention. Various hyperparameter ranges can be explored, including num_leaves, max_depth, learning_rate, n_estimators, subsample_for_bin, and class_weight.

*f. Calibrating Output Probabilities:*

To address the bias in predicted probabilities from some machine learning algorithms, the following calibration approaches were considered:

- Calibration using mean and standard deviation of historical attrition data by various data cuts.
- Calibration using sigmoid (parametric) calibration, assuming calibrated predicted probabilities follow a sigmoid shape.
- Calibration using isotonic (non-parametric) calibration, which does not assume any specific shape for the calibration probabilities.

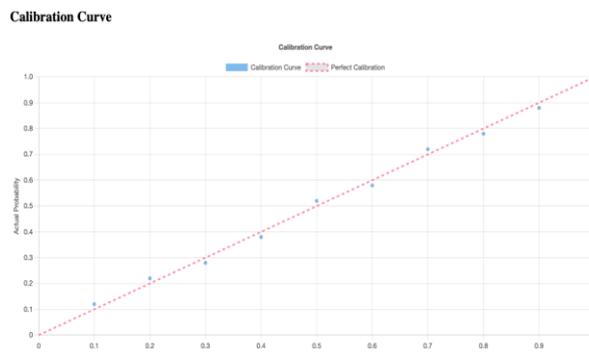

By carefully considering and evaluating these design choices, the employee attrition model can provide accurate and interpretable predictions while addressing challenges such as imbalanced data, categorical features, and the incorporation of employee sentiment data.

## 6. CONCLUSION

The data-driven approach to mitigating employee attrition presented in this paper offers a powerful tool for organizations to address one of their most pressing challenges in talent management. By leveraging machine learning algorithms and comprehensive data engineering practices, this framework provides a proactive method for identifying and addressing attrition risks before they manifest.

The proposed methodology successfully tackles several key challenges in attrition prediction, including the handling of imbalanced datasets, incorporation of diverse data sources, and interpretation of complex models. The emphasis on feature engineering ensures that the model captures a wide array of factors influencing employee decisions, while the focus on model interpretation through techniques like SHAP values enables organizations to understand the drivers of attrition and develop targeted retention strategies.

The design choices discussed, from algorithm selection to probability calibration, provide a robust foundation for building effective attrition prediction models. The use of advanced techniques such as LightGBM and isotonic calibration addresses the complexities inherent in HR data, resulting in more accurate and reliable predictions. This approach represents a significant advancement in the field of HR analytics, moving beyond traditional reactive measures to a more sophisticated, data-driven strategy. By enabling organizations to predict attrition risks at both individual and aggregate levels, it allows for more efficient resource allocation and personalized retention efforts.

However, it is important to note that the effectiveness of any such system depends on the quality and comp-ehensiveness of the data available, as well as the organization's commitment to acting on the insights generated. Future research could explore the long-term impact of implementing such systems on organizational performance, employee satisfaction, and overall retention rates. In conclusion, this data-driven approach to attrition mitigation offers organizations a powerful tool to enhance their talent management strategies, reduce costs associated with turnover, and maintain a competitive edge in today's dynamic business environment. As the field of HR analytics continues to evolve, such predictive models will likely become an essential component of strategic workforce planning and management.